\documentclass[graybox]{svmult}

\usepackage{type1cm}        
\usepackage{makeidx}         
\usepackage{graphicx}        
\usepackage{multicol}        
\usepackage[bottom]{footmisc}
\usepackage{url}
\usepackage[numbers]{natbib}
\usepackage{algorithm, eqparbox, array} 
\usepackage{algcompatible}
\usepackage{algpseudocode}
\usepackage{booktabs}
\usepackage{subcaption}
\usepackage{newtxtext}      
\usepackage[varvw]{newtxmath}       

\makeindex             

\usepackage{todonotes}

\begin{document}

\title*{Handling Concept Drift in Global Time Series Forecasting}
\author{Ziyi Liu, Rakshitha Godahewa, Kasun Bandara and Christoph Bergmeir}
\institute{Ziyi Liu \at Department of Data Science and AI, Monash University, Melbourne, Australia \\ \email{ziyi.liu1@monash.edu}
\and Rakshitha Godahewa \at Department of Data Science and AI, Monash University, Melbourne, Australia \\ \email{rakshitha.godahewa@monash.edu}
\and Kasun Bandara \at School of Computing and Information Systems, University of Melbourne, Melbourne, Australia 
\at EnergyAustralia, Melbourne, Australia \\
\email{kasun.bandara@unimelb.edu.au }
\and Christoph Bergmeir (corresponding author) \at Department of Data Science and AI, Monash University, Melbourne, Australia \\ \email{christoph.bergmeir@monash.edu}}

\maketitle

\abstract{Machine learning (ML) based time series forecasting models often require and assume certain degrees of stationarity in the data when producing forecasts. However, in many real-world situations, the data distributions are not stationary and they can change over time while reducing the accuracy of the forecasting models, which in the ML literature is known as \textit{concept drift}. Handling concept drift in forecasting is essential for many ML methods in use nowadays, however, the prior work only proposes methods to handle concept drift in the classification domain. To fill this gap, we explore concept drift handling methods in particular for Global Forecasting Models (GFM) which recently have gained popularity in the forecasting domain. We propose two new concept drift handling methods, namely: Error Contribution Weighting (ECW) and Gradient Descent Weighting (GDW), based on a continuous adaptive weighting concept. These methods use two forecasting models which are separately trained with the most recent series and all series, and finally, the weighted average of the forecasts provided by the two models are considered as the final forecasts. Using LightGBM as the underlying base learner, in our evaluation on three simulated datasets, the proposed models achieve significantly higher accuracy than a set of statistical benchmarks and LightGBM baselines across four evaluation metrics.}

\section{Introduction}
\label{sec:introduction}

Accurate time series forecasting is crucial in the context of decision making and strategic planning nowadays for many businesses and industries. The forecasting models are typically trained based on historical data. In most applications, data distributions are not stationary and they change over time while making the trained models outdated and reducing their forecasting accuracy, which is known as \emph{concept drift}. Hence, it is important to make sure that the trained models well-generalise beyond the training data and are capable to provide accurate forecasts even with the changed data distributions. Transfer learning can be applied to make the models robust to distribution shift. In general, transfer learning is an approach of solving one problem and applying the corresponding knowledge gained to solve another related problem. In the time series forecasting domain, the transfer learning models typically train on a large set of series and during forecasting, they predict the future of (1) new series that they have not seen before; (2) training series where the data distributions change over time. In this chapter, we focus on the second sub-problem of transfer learning which is predicting the future of a set of series that is used to train a model when the corresponding data distribution is not stationary.

Traditional forecasting methods like Exponential Smoothing  \citep[ETS,][]{ref_112} are capable of handling the most common non-stationarities, namely trends and seasonalities, and also have been designed to show a certain degree of robustness under other types of non-stationarities. In particular, ETS gives most weight to the most recent observations and therefore it can adapt to a distribution shift reasonably well. However, Global Forecasting Models \citep[GFM,][]{ref_105} have recently shown their potential in providing more accurate forecasts compared to the traditional univariate forecasting models, by winning the M4 \citep{ref_30} and M5 \citep{makridakis_2020_m5} forecasting competitions. In contrast to traditional univariate forecasting models that build isolated models on each series, GFMs build a single model across many series while allowing the model to learn the cross-series information. By building a single model over many series, the models have more data available for training and can afford to be more complex \cite{pablo_2020_principles}. Consequently, ML methods like neural networks and Gradient Boosted Trees are usually the methods of choice for global modelling. However, unlike many traditional forecasting methods like ETS, ML methods do normally not have inherent mechanisms to deal with changes in the underlying distribution. Therefore, in ML the field of concept drift has emerged, which studies how ML methods can be used under distribution shift \cite{webb2016characterizing}. In this chapter, we study how the methods from the concept drift literature can be used for forecasting with ML methods.

To the best of our knowledge, no methods to deal with concept drift in the forecasting domain in a GFM and ML context have been proposed in the literature. The available concept drift handling methods are all related to the classification domain. Dynamic Weighted Majority \citep[DWM,][]{kolter2007dynamic},  Social Adaptive Ensemble \citep[SAE,][]{gomes2013sae} and Error Interaction Approach \citep[EIA,][]{baier2020handling} are some of the popular concept drift handling methods in the classification domain. Most of these concept drift handling methods use ensembling \citep{gomes2017survey}. In the forecasting space, ensembling is also known as forecast combination which aggregates the predictions of multiple forecasting models to produce the final forecasts. Ensembling techniques are widely used to reduce model variance and model bias. This motivates us to explore methods to handle concept drift in the GFM space that incorporate ensembling techniques.

In this study, we propose two new forecast combination methods to handle the concept drift in the forecasting domain. We name these two methods Error Contribution Weighting (ECW) and Gradient Descent Weighting (GDW). In particular, these methods independently train two forecasting models based on different time series, typically one model is trained with the most recent series history and the other model is trained with the full series history, where the two models are finally weighted based on the previous prediction error, which is known as continuous adaptive weighting. The proposed ECW and GDW methods are different based on the method of calculation of the sub-model weights. In the proposed methods, the sub-model trained with the full series' history is expected to provide more accurate forecasts in general situations whereas if concept drift occurs, then the sub-model trained with the most recent series' history is expected to quickly adapt to the new concept. Furthermore, two different methods for series weighting, exponential weighting and linear weighting, are used during the training of the two sub-models where these methods internally assign higher weights for the most recent series observations than the older observations during training. We also quantify the effect of handling concept drift using our proposed methods and only using the series weighted methods considering LightGBM \citep{guolin_lightgbm_2017} as the base learner. The LightGBM models that use our proposed concept drift handling methods outperform a set of statistical benchmarks and LightGBM baselines with statistical significance across three simulated datasets on four error metrics. All implementations of this study are publicly available at: \url{https://github.com/Neal-Liu-Ziyi/Concept_Drift_Handling}.

The remainder of this chapter is organized as follows: Section \ref{sec:problem} introduces different concept drift types. Section \ref{sec:related_work} reviews the relevant prior work. Section \ref{sec:methodology} explains the series weighted methods and proposed concept drift handling methods in detail. Section \ref{sec:experiments} explains the experimental framework, including the datasets, error metrics, benchmarks, evaluation and statistical testing. An analysis of the results is presented in Section \ref{sec:results}. Section \ref{sec:concluson} concludes the study and discusses possible future research.

\section{Problem Statement: Concept Drift Types}
\label{sec:problem}
There are two main types of concept drift: real concept drift \citep{kolter2007dynamic} and virtual concept drift \citep{delany2004case}. Real concept drift occurs when the true outputs of the instances change over time whereas virtual concept drift occurs when the data distribution changes over time even with the same true outputs, possibly due to noise. In many situations, real and virtual concept drift can co-occur. 

\begin{figure}[!htbp]
    \centering
    \includegraphics[width=0.6\textwidth]{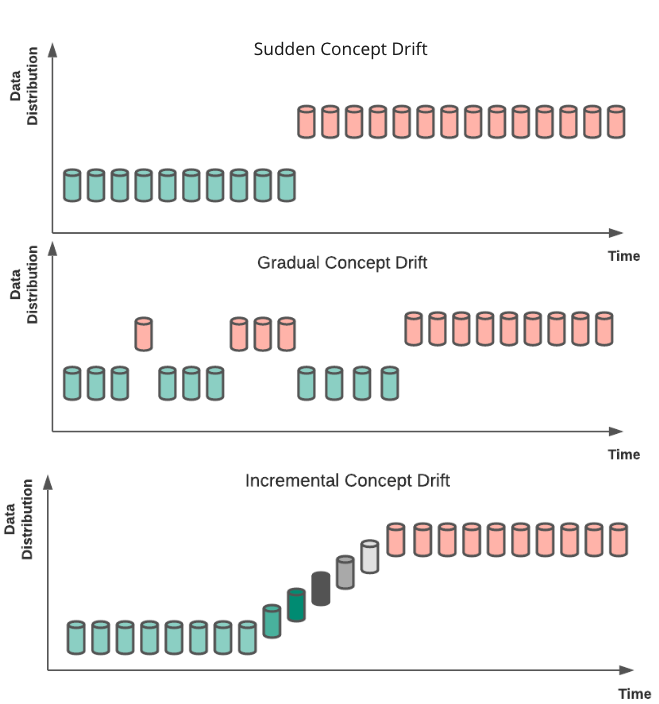}
    \caption{A visualisation of sudden, gradual and incremental concept drift types.}
    \label{fig: 3_cd}
\end{figure}

The literature typically classifies real concept drift into four different groups \cite{webb2016characterizing} which are (1) sudden concept drift where the data distribution at time $t$ suddenly changes into a new distribution at time $t+1$; (2) incremental concept drift where the data distribution changes and stays in a new distribution after going through a variable distribution; (3) gradual concept drift where the data distribution shifts between a new distribution and the current distribution over time where gradually, the series completely enters the new distribution; (4) recurring concept drift where the same concept drift occurs periodically. Incremental and recurring concept drift in a time series forecasting context are closely related to trends and seasonalities in the data. In particular, recurring concept drift can be easily handled by the typical handling of seasonality with dummy variables, Fourier terms and similar additional features \cite{ref_4}. Thus, in this study we focus on three concept drift types: sudden, incremental and gradual, which are illustrated in Figure \ref{fig: 3_cd}. 

\section{Related Work}
\label{sec:related_work}

In the following, we discuss the related prior work in the areas of global time series forecasting, transfer learning and concept drift handling. 

\subsection{Global Time Series Forecasting}
\label{sec: rw_gfm}

GFMs \citep{ref_105} build a single model across many series with a set of global parameters that are the same across all series. In contrast to the traditional univariate forecasting models, GFMs are capable of learning cross-series information with a fewer amount of parameters. Global models have been pioneered by works such as \citet{ref_99, ref_1, ref_2, ref_6, godahewa_2021_ensembling}.

GFMs have recently obtained massive popularity in the forecasting field after winning the M4 \citep{ref_30} and M5 \citep{makridakis_2020_m5} forecasting competitions. In particular, the winning method of the M4 competition uses global Recurrent Neural Networks \citep[RNN,][]{ref_6} whereas the winning method of the M5 competition uses global LightGBM \cite{guolin_lightgbm_2017} models. 

\subsection{Transfer Learning}
\label{sec: transfer_learning}
In the forecasting domain, there are many works that present on forecasting new time series that a model has not seen during training. \citet{oreshkin_2021_meta} propose a deep learning based meta-learning framework that generalises well on new time series from different datasets that it has not seen before. The popular Nixtla framework \citep{mergenthaler_2022_nixtla} provides ten N-BEATS \citep{oreshkin_2019_nbeats} and N-HiTS \citep{challu_2023_nhits} models that are pre-trained using the M4 dataset \citep{ref_30} where the models are expected to generalise on unseen time series. \citet{woo_2022_deeptime} propose DeepTime, a meta-learning framework that can forecast unseen time series by properly addressing distribution shifts using a ridge regressor. \citet{grazzi_2021_meta} propose Meta-GLAR, which performs transfer learning using local closed-form adaptation updates. The transfer learning framework proposed by \citet{ref_106} in particular considers the adequate long-ago data during training. 

To the best of our knowledge, all these transfer learning frameworks produce forecasts for unseen time series. In contrast, our proposed methods produce forecasts for the same set of time series used during training where the distribution of the series is not stationary.

\subsection{Concept Drift Handling}
\label{sec: rw_concept_drift}

To the best of our knowledge, the existing concept drift handling methods are all related to the classification domain. The concept drift handling methods in the literature can be mainly divided into two categories: implicit methods and explicit methods \citep{ghomeshi2019eacd}. The explicit methods use a drift detection technique and immediately react to the drift at the point of its occurrence. The implicit methods do not use drift detection techniques. 

In particular, \citet{chu2004fast} propose the Adaptive Boosting Model which is an explicit method of concept drift handling in the classification domain. This method assigns a weight to each training instance where the weights are consequently updated across a sequence of classifiers. The first classifier assigns a sample weight to each instance. Each classifier then keeps the same weights for the correctly classified instances and assigns higher weights for the incorrectly classified instances where the weighted instances are used to train the next classifier in the sequence. Each classifier is also assigned a separate weight based on its classification accuracy. This process is repeated until the error function does not change. The weighted average of the predictions provided by all classifiers are considered as the final prediction. When a concept drift situation is detected, the weight of each classifier in the sequence is reset to one. However, this method contains a few limitations. It is highly time-consuming for trivial concept drift. Also, if the concept drift is too small to impact the accuracy of the previous classifier, it may stop building new classifiers. Furthermore, it needs to reset all classifiers when a data stream encounters concept drift which may then lead the model to be sensitive to false alarms \citep{krawczyk2017ensemble}. The recurring concept drift framework, RCD, proposed by \citet{gonccalves2013rcd} also uses an explicit approach to handle concept drift. This method extends a similar approach proposed by \citet{widmer1996learning} that was originally designed to deal with categorical data, to work with numerical data. It stores the context of each data type in a buffer and a classifier is built together with a concept drift detector. This method uses a two step drift detection mechanism. The drift detector uses a new classifier to detect concept drift. If there is no concept drift, an empty buffer is filled with the samples used in the training of the classifier. Furthermore, there is a warning level to monitor the error rate of the classifier where if the error rate of the classifier reaches the warning level, it means concept drift might be occurring. Simultaneously, a new classifier is created with the new buffer. If the error rate of this classifier decreases and returns to the normal level, then the algorithm will treat that concept drift as a false alarm. Furthermore, when a true concept drift occurs, a statistical test compares the current buffer with all previous buffers to detect whether the new concept has already been observed in the past. If the test result is positive, which means the new concept has been seen before, then the previous classifier and the buffer corresponding to this concept are used. Otherwise, the current classifier and the buffer are stored in the list.

The main limitation of these explicit methods is that they frequently falsely detect concept drift. Therefore, using these explicit methods may result in incorrect concept drift detection while reducing the model performance. Furthermore, it is difficult to use a drift detector to recognise different concept drift types. Thus, implicit methods are usually more useful in handling concept drift. 

The DWM method proposed by \citet{kolter2007dynamic} is an implicit method of handling concept drift. This method uses a four-step process to handle concept drift: (1) the DWM holds a weighted pool of base learners which adds or removes base learners based on the performance of the global algorithm; (2) if the global algorithm makes a mistake, then an expert will be added; (3) if a base learner makes a mistake, then its weight will be reduced; (4) if the weight of a base learner reduces below a threshold, then it will be removed. The SAE method proposed by \citet{gomes2013sae} is also a popular implicit method which is also known as the social network abstraction for data stream classification. It uses a similar learning strategy  as the DWM method. The SAE method explores the similarities between its classifiers. It also drops classifiers based on their accuracy and similarities with other classifiers. The new classifiers are only added if the overall method accuracy is lower than a given threshold. Furthermore, SAE connects two classifiers when they have very similar predictions.

Even though implicit methods overcome the issues of explicit methods such as the model sensitivity to falsely detecting concept drift and difficulties in recognising different concept drift types, they take a considerable amount of time to adapt to a new concept. This phenomenon has motivated researchers to explore robust and rapidly reactive concept drift handling methods without drift detection. 

The EIA method proposed by \citet{baier2020handling} is an implicit method that does not use drift detection. The EIA uses two models with different complexities, $M_{simple}$ and $M_{complex}$, to make predictions. At a time, the predictions are obtained using a single model where this model is selected based on its error predicted using a regression method. Here, $M_{complex}$ captures more information during model training and thus, in normal situations, $M_{complex}$ is used to make predictions. However, when the data has concept drift, in particular sudden concept drift, $M_{complex}$ is unable to respond to the concept drift quickly. Thus, in such situations, $M_{simple}$ which is trained with most recent observations is applied as it can quickly adapt to the current concept. However, the EIA method is only applicable to sudden concept drift as it is difficult to switch between the two models more frequently to handle continuous drift types such as incremental and gradual concept drift types.

In line with that, in this study, we propose two methods which use the continuous adaptive weighting approach, ECW and GDW, where both methods are inspired by the EIA method. Similar to the EIA, our proposed methods use two sub-models. However, unlike the EIA method, the weighted average of the forecasts provided by both sub-models is considered during forecasting where the weights of the sub-models are dynamically changed. Furthermore, all the above explained concept drift handling work is from the classification domain where in this study, we explore the elements of these methods that should be changed to adapt them to handle concept drift in the forecasting domain.

\section{Methodology}
\label{sec:methodology}
This section explains the series weighted methods and our proposed concept drift handling methods in detail.

\subsection{Series Weighted Methods}
\label{sec:series_weighted_methods}

In this simple and straightforward first approach, we assign different weights for the training instances. In particular, we consider the most recent instances to be more relevant than older instances and thus, during training, we assign higher weights  for the most recent instances where the weights are gradually decreased for the older instances, e.g., using an exponential decay similar to ETS.

In our work, we consider two series weighting methods: exponential weighting and linear weighting. Equations \ref{eqn:swm_1} and \ref{eqn:swm_2} respectively define the process of weighting the training instances and the method of calculating the weights with exponential and linear weighting methods. Here, $x_i$ is the original $i^{th}$ instance, $X_i$ is the weighted $i^{th}$ instance, $\text{series\_length}$ is the number of training instances  considered from a particular series, $\alpha_{\text{series\_length}-i}$ is the weight of the $i^{th}$ instance, $\alpha_0$ is the initial weight of the most recent instance, where $0< \alpha \leq 1 $, and $\beta$ is the weight decrease of the linear weighting, where $0<\beta \leq 1$.

\begin{equation} 
\label{eqn:swm_1}
    X_i = \alpha_{\text{series\_length}-i} \cdot x_i
\end{equation}

\begin{equation} 
\label{eqn:swm_2}
     \alpha_i =\begin{cases}
       \alpha_{i-1}\cdot \alpha_0 &\text{exponential weighting} \\
       \alpha_{i-1}-\frac{\beta}{\text{series\_length}} &\text{linear weighting} 
        \end{cases}\\
\end{equation}

The experiments are conducted considering two values for $\text{series\_length}$, 200 instances and all instances. The values of $\alpha_0$ and $\beta$ are fixed to 0.9 based on our preliminary experiments. Finally, four models are trained with each of our experimental datasets (Section \ref{sec:datasets}), considering the two weighting methods: exponential and linear weighting, and two $\text{series\_length}$ values.

\subsection{Our Proposed Concept Drift Handling Methods}
\label{sec:continuous_adaptive_weighting}
We propose two continuous adaptive weighting methods: ECW and GDW. In particular, our methods train two forecasting models with different series where one model is trained with the recent series observations ($M_{\textit{partial}}$) and the other model is trained with all series observations ($M_{\textit{all}}$). The weighted average of the predictions provided by $M_{\textit{partial}}$ and $M_{\textit{all}}$ is considered as the final prediction of a given time point where the weights of these models are dynamically changed and calculated based on the previous prediction error. As $M_{\textit{partial}}$ is trained with the recent data, its prediction accuracy may be less than the prediction accuracy of $M_{\textit{all}}$, however, if the series encounters concept drift, then $M_{\textit{partial}}$ can quickly adapt to the new concept compared to $M_{\textit{all}}$. In general, the weight of $M_{\textit{partial}}$ is less than the weight of $M_{\textit{all}}$ due to its lower prediction accuracy. However, when a concept drift occurs, our methods rapidly increase the weight of $M_{\textit{partial}}$ which can efficiently remedy the issues of other implicit concept drift handling methods.

We consider Residual Sum of Squares (RSS) as the loss function of $M_{\textit{partial}}$ and $M_{\textit{all}}$. Equation \ref{eqn:rss} defines RSS where $y_i$ are the actual values, $\hat y_i$ are the predictions and $n$ is the number of predictions.

\begin{equation}
\label{eqn:rss}
	RSS = \sum_{i=1}^n (y_i-\hat y_i)^2
\end{equation}

As we only consider the error of the last prediction for the weight calculations corresponding with the next prediction, Equation \ref{eqn:rss} can be reduced to:

\begin{equation}
	RSS_i = (y_{i}-\hat y_{i})^2
\end{equation}

Based on preliminary experiments and to manage complexity of the experiments, we consider 200 time series observations when training the $M_{\textit{partial}}$ model. 

Our proposed ECW and GDW methods are different based on the algorithms they use to calculate the weights of $M_{\textit{partial}}$ and $M_{\textit{all}}$, which we explain in detail in the next sections.

\subsubsection{Error Contribution Weighting}
\label{sec:ecw}
\index{Error Contribution Weighting}
The ECW method directly outputs the prediction of $M_{\textit{all}}$ as the first prediction. From the second prediction onwards, it determines the weights that should be used to combine the predictions of the two sub-models as follows.

The prediction errors of $M_{\textit{partial}}$ and $M_{\textit{all}}$ for the $i^{th}$ prediction, $\epsilon_{p_i}$ and  $\epsilon_{a_i}$, are defined as shown in Equations \ref{eqn: ep_pi} and \ref{eqn: ep_ai}, respectively, considering RSS as the loss function. 

\begin{equation}
\label{eqn: ep_pi}
\epsilon_{p_i}= (y_i-\hat y_{\textit{partial}_i})^2
\end{equation}

\begin{equation}
\label{eqn: ep_ai}
\epsilon_{a_i}= (y_i-\hat y_{\textit{all}_i})^2
\end{equation}

The error percentages of $M_{\textit{partial}}$ and $M_{\textit{all}}$ for the $i^{th}$ prediction, $E_{p_i}$ and  $E_{a_i}$, are defined as shown in Equations \ref{eqn: epp_pi} and \ref{eqn: epp_ai}, respectively. 
\begin{equation}
\label{eqn: epp_pi}
E_{p_{i}}=\frac{\epsilon_{p_i}}{\epsilon_{p_i}+\epsilon_{a_i}}\\ 
\end{equation}

\begin{equation}
\label{eqn: epp_ai}
E_{a_{i}}=\frac{\epsilon_{a_i}}{\epsilon_{p_i}+\epsilon_{a_i}}
\end{equation}

The weights corresponding with $M_{\textit{partial}}$ and $M_{\textit{all}}$ for the $i^{th}$ prediction, $w_{p_i}$ and  $w_{a_i}$, can be then defined as shown in Equations \ref{eqn: om_pi} and \ref{eqn: om_ai}, respectively. In this way, the higher weight is assigned to the model with the lowest previous prediction error.

\begin{equation}
\label{eqn: om_pi}
w_{p_{i}}=E_{a_{i-1}}
\end{equation}

\begin{equation}
\label{eqn: om_ai}
w_{a_{i}}=E_{p_{i-1}}
\end{equation}

Equation \ref{eqn: ecw_p} shows the formula used to obtain the final prediction corresponding with the $i^{th}$ time point, $\hat y_i$, after weighting. 

\begin{equation}
\label{eqn: ecw_p}
\hat y_i=w_{p_{i}} \cdot \hat y_{partial_i}+ w_{a_{i}} \cdot \hat y_{all_i}
\end{equation}

Algorithm \ref{alg:ECW_alg} shows the process of the ECW method that is used to obtain the prediction for a given time point. 

\begin{algorithm}
\begin{algorithmic}[1] 
\Procedure{ecw}{$y_{i-1}$, $\hat y_{partial_{i-1}}$, $\hat y_{all_{i-1}}$, $\hat y_{partial_i}$, $\hat y_{all_i}$, $i$} 
\If{$i$ is 1}
\State $y_i \gets \hat y_{all_i}$
\Else 
\State $\epsilon_{a_{i-1}} \gets calculate\_rss(y_{i-1},\hat y_{all_{i-1}})$
\State $\epsilon_{p_{i-1}} \gets  calculate\_rss(y_{i-1},\hat y_{partial_{i-1}})$
\State $w_{a_{i}} \gets \frac{\epsilon_{p_{i-1}}}{\epsilon_{p_{i-1}} + \epsilon_{a_{i-1}}}$
\State $w_{p_{i}} \gets \frac{\epsilon_{a_{i-1}}}{\epsilon_{p_{i-1}} +\epsilon_{a_{i-1}}}$
\State $y_i \gets w_{p_{i}} \cdot \hat y_{partial_i}+ w_{a_{i}} \cdot \hat y_{all_i}$
\EndIf 
\State \Return $y_i$
\EndProcedure
\caption{Error Contribution Weighting}
\label{alg:ECW_alg}
\end{algorithmic}
\end{algorithm}

\subsubsection{Gradient Descent Weighting}
\label{sec:gdw}
\index{Gradient Descent Weighting}
The GDW method uses gradient descent to determine the weights of the sub-models. The sub-models are initialised with weights of $0.5$. This method also directly outputs the prediction of $M_{\textit{all}}$ as the first prediction. From the second prediction onwards, it determines the weights that should be used to combine the predictions of the two sub-models using gradient descent as follows.

The final error of the $i^{th}$ prediction, $\epsilon_{i}$, is defined as shown in Equation \ref{eqn: ep_i} considering RSS as the loss function. 

\begin{equation}
\label{eqn: ep_i}
\epsilon_i=(y_i-w_{p_i}\cdot \hat y_{\textit{partial}_i}-w_{a_i}\cdot \hat y_{\textit{all}_i})^2
\end{equation}

The gradients calculated from $M_{\textit{partial}}$ and $M_{\textit{all}}$ for the $i^{th}$ prediction, $g_{{p_i}}$ and  $g_{{a_i}}$, are defined as shown in Equations \ref{eqn: g_pi} and \ref{eqn: g_ai}, respectively. 

\begin{equation}
\label{eqn: g_pi}
g_{p_i}=\nabla \epsilon_i(w_{p_i})=-2\cdot \hat y_{\textit{partial}_i}\cdot \epsilon_i
\end{equation}

\begin{equation}
\label{eqn: g_ai}
g_{a_i}=\nabla \epsilon_i(w_{a_i})=-2\cdot \hat y_{\textit{all}_i}\cdot \epsilon_i
\end{equation}

To avoid large gaps between two adjacent weights, the gradients are multiplied by an adjusted rate $\eta$. The updated weights of $M_{\textit{partial}}$ and $M_{\textit{all}}$ for the $i^{th}$ prediction, $w_{p_i}$ and $w_{a_i}$, are then defined as shown in Equations \ref{eqn: og_pi} and \ref{eqn: og_ai}, respectively. Based on our preliminary experiments, the value of $\eta$ is fixed to 0.01.

\begin{equation}
\label{eqn: og_pi}
w_{p_i} = w_{p_{i-1}}-g_{p_{i-1}}\cdot \eta\\
\end{equation}

\begin{equation}
\label{eqn: og_ai}
w_{a_i} = w_{a_{i-1}}-g_{a_{i-1}}\cdot \eta
\end{equation}

Equation \ref{eqn: gdw_p} shows the formula used to obtain the final prediction corresponding with the $i^{th}$ time point after weighting.

\begin{equation}
\label{eqn: gdw_p}
\hat y_i=w_{p_i} \cdot \hat y_{partial_i}+ w_{a_i} \cdot \hat y_{all_i}
\end{equation}

Algorithm \ref{alg:EGDW_alg} shows the process of the GDW method that is used to obtain the prediction for a given time point. 

\begin{algorithm}
\begin{algorithmic}[1]
\Procedure{gdw}{$y_{i-1}$, $\hat y_{i-1}$, $\hat y_{partial_{i-1}}$, $\hat y_{all_{i-1}}$, $\hat y_{partial_i}$, $\hat y_{all_i}$, $w_{p_{i-1}}$, $w_{a_{i-1}}$, $\eta$, $i$} 
\If{$i$ is 1}
\State $y_i \gets \hat{y}_{all_i}$
\Else 
\State $\epsilon_{i-1} \gets calculate\_rss(y_{i-1},\hat y_{i-1})$
\State $g_{p_{i-1}} \gets  -2\cdot \hat y_{partial_{i-1}}\cdot \epsilon_{i-1}$
\State $g_{a_{i-1}} \gets  -2\cdot \hat y_{all_{i-1}}\cdot \epsilon_{i-1}$
\State $w_{p_i} =w_{p_{i-1}}-g_{{p}_{i-1}}\cdot \eta$
\State $w_{a_i}=w_{a_{i-1}}-g_{{a}_{i-1}}\cdot \eta$
\State $y_i \gets w_{p_{i}} \cdot \hat y_{partial_i}+ w_{a_{i}} \cdot \hat y_{all_i}$
\EndIf
\State \Return $y_i$
\EndProcedure
\caption{Gradient Descent Weighting}
\label{alg:EGDW_alg}
\end{algorithmic}
\end{algorithm}

\bigskip

When training $M_{\textit{partial}}$ and $M_{\textit{all}}$ models, we consider both exponential and linear weighting methods which internally weight the training instances as explained in Section \ref{sec:series_weighted_methods}. Thus, the forecasts are obtained from four model combinations, considering two models and two weighting methods. The final forecasts of the ECW and GDW methods are obtained by averaging the forecasts provided by the above four model combinations.

Our proposed methods are applicable to any GFM including neural networks, machine learning and deep learning models. However, for our experiments, we consider LightGBM \citep{guolin_lightgbm_2017} as the base learner due to its recent competitive performance over the state-of-the-art forecasting models including deep learning models \citep{JANUSCHOWSKI2021}.

\section{Experimental Framework}
\label{sec:experiments}
In this section, we discuss the experimental datasets, error metrics, evaluation method, and benchmarks used in our experiments.

\subsection{Datasets}
\label{sec:datasets}

We are unaware of any publicly available real-world datasets which contain series showing sudden, incremental and gradual concept drift types, adequately, to evaluate the performance of our proposed methods, ECW and GDW. Thus, in this study, we limit ourselves to the use of simulated datasets that contain series representing sudden, incremental and gradual concept drift types. 

Let $\textit{ts}_1$ and $\textit{ts}_2$ be two time series simulated from the same AR(3) data generation process with different initial values and a random seed. The two series are then combined based on the required concept drift type as shown in Equation \ref{eq:simulation}, where $x_i$ is the $i^{th}$ element of the combined series and $\text{series\_length}$ is the length of the series. 

\begin{equation}
    \begin{split}   x_i (\textit{sudden}) &=\begin{cases}
       \textit{ts}_{1_i} &\text{if } i <t_{\textit{drift}} \\
       \textit{ts}_{2_i} &\text{if } i \geq t_{\textit{drift}}
        \end{cases}\\
         x_i (\textit{incremental})  &=\begin{cases}
       \textit{ts}_{1_i} &\text{if } i <t_{\textit{start}} \\
       (1-w)\textit{ts}_{1_i}+(w)\textit{ts}_{2_i} &\text{if } t_{\textit{start}} \leq i\leq t_{\textit{end}}\\
       \textit{ts}_{2_i} &\text{if } i > t_{\textit{end}}
        \end{cases}\\
    x_i (\textit{gradual}) &=\begin{cases}
       \textit{ts}_{1_i} &\text{if } I_i=0 \\
       \textit{ts}_{2_i} &\text{if } I_i=1
        \end{cases}\\
    \end{split}
    \label{eq:simulation}
\end{equation}

\text{Where } $$w=\frac{i-t_{\textit{start}}}{t_{\textit{start}}-t_{\textit{end}}}$$
$I_i\  \text{is random based on probability }p_i$
$$p_i=\frac{i}{\text{series\_length}}$$

For sudden concept drift, the drift point, $t_{\textit{drift}}$, is randomly selected. The combined series contains $\textit{ts}_1$ before $t_{\textit{drift}}$ and $\textit{ts}_2$ on and after $t_{\textit{drift}}$. For incremental concept drift, the drift starting point, $t_{\textit{start}}$, and drift ending point, $t_{\textit{end}}$, are randomly selected. The combined series contains $\textit{ts}_1$ before $t_{\textit{start}}$ and $\textit{ts}_2$ after $t_{\textit{end}}$. A linear weighted combination of $\textit{ts}_1$ and $\textit{ts}_2$ is considered in between $t_{\textit{start}}$ and $t_{\textit{end}}$ where $w$ and $(1-w)$ are the weights of $\textit{ts}_2$ and $\textit{ts}_1$, respectively. For gradual concept drift, the combined series contains values from either $\textit{ts}_1$ or $\textit{ts}_2$ depending on the corresponding $I_i$ which is randomised based on an increasing probability, $p_i$. In general, $p_i$ is used to represent the probability that an instance belongs to $\textit{ts}_2$. Thus, when $p_i=1$ the series ultimately enters the new distribution. 
Finally, three simulated datasets are created that respectively contain series representing sudden, incremental and gradual concept drift types. Each dataset has 2000 series. The length of each series in all datasets is 2000. The first 1650 data points in each series are used for model training whereas the last 350 data points in each series are reserved for testing. Note that the drift can occur in either training or testing parts of the series.
 
\subsection{Error Metrics}
\label{sec:error_metrics}

We measure the performance of our models using Root Mean Squared Error (RMSE) and Mean Absolute Error \citep[MAE,][]{sammut_2010_mae} which are commonly used error metrics in the time series forecasting field. Equations \ref{eqn:rmse} and \ref{eqn:mae} respectively define the RMSE and MAE error metrics. Here, $F_k$ are the forecasts, $Y_k$ are the actual values for the required forecast horizon and $h$ is the forecast horizon.

\begin{equation}
\label{eqn:rmse}
    RMSE = \sqrt{\frac{\sum_{k=1}^{h}{|F_{k} - Y_{k}|}^2}{h}}
\end{equation}

\begin{equation}
\label{eqn:mae}
    MAE = \frac{\sum_{k=1}^{h}{|F_{k} - Y_{k}|}}{h}
\end{equation}

Since all these error measures are defined for each time series, we calculate the mean and median values of them across a dataset to measure the model performance. Therefore, four error metrics are used to evaluate each model: mean RMSE, median RMSE, mean MAE and median MAE.

\subsection{Evaluation}
\label{sec:prequential_evaluation}

We use prequential evaluation \citep{dawid1984present} to evaluate the performance of the models. This procedure is also a common evaluation technique used in the forecasting space, where it is normally called time series cross-validation \cite{ref_4}. We conduct the prequential evaluation with increasing block sizes \citep{cerqueira2020evaluating} as follows. 

First, the forecast horizon of each series which consists of 350 data points is divided into 7 test sets of 50 data points each. The models are then trained with the training set and 50 predictions are obtained corresponding with the first test set, one at a time iteratively. This procedure is also known as rolling origin without recalibration \cite{tashman2000out} as the models are not retrained when obtaining individual predictions. After obtaining the predictions corresponding with the first test set, the actuals corresponding with the first test set are also added to the training set. The models are retrained with the new training set and the 50 predictions corresponding with the second test set are obtained, one at a time iteratively, in a rolling origin with recalibration fashion. This process is repeated until all 350 predictions are obtained corresponding with the complete forecast horizon. 

\subsection{Benchmarks}
\label{sec:benchmarks}

We consider three types of LightGBM models: plain LightGBM that does not handle concept drift (Plain), LightGBM trained with exponential weighting (EXP) and LightGBM trained with linear weighting (Linear), as the main benchmarks of our study. We also consider three statistical models: AR(3) model, AR(5) model and ETS \citep{ref_112} as the benchmarks of our study. All statistical models are implemented using the Python package, \verb|StatsForecast| \citep{garza2022statsforecast}.

Each baseline is separately trained considering all data points and the last 200 data points in the training series. Thus, twelve models are finally considered as benchmarks, considering six LightGBM models and six statistical models.  

\subsection{Statistical Testing of the Results}
\label{sec: statistical_testing}
The non-parametric Friedman rank-sum test is used to assess the statistical significance of the results provided by different forecasting models across time series considering a significance level of $\alpha$ = 0.05. Based on the corresponding RMSE errors, the methods are ranked on every series of a given experimental dataset. The best method according to the average rank is chosen as the control method. To further characterise the statistical differences, Hochberg's post-hoc procedure \citep{garcia2010advanced} is used.

\section{Results and Discussion}
\label{sec:results}
This section provides a comprehensive analysis of the results of all considered models in terms of main accuracy and statistical significance, and later also gives more insights into the modelling.

\subsection{Main Accuracy Results}

Table \ref{tab:all_results} shows the results of all models across sudden, incremental and gradual concept drift types for mean RMSE, median RMSE, mean MAE and median MAE. 

\begin{table}
		\centering
		\caption{Results across sudden, incremental and gradual concept drift types. The best performing models in each group are italicized and the overall best performing models are highlighted in boldface.}
		\begin{tabular*}{\textwidth}{l @{\extracolsep{\fill}} lrrrr}
			\toprule
			&  & Mean RMSE & Median RMSE  & Mean MAE  & Median MAE\\
			\cmidrule{3-6}
			\addlinespace
			\multicolumn{5}{l}{\bf Sudden} \\
			\addlinespace
			 & AR3\_{200}  &     0.7318    &      0.5154 &   0.6184 & 
            0.4933\\
            & AR3\_{All}  &      0.5973  &     0.5303     &   \textbf{\textit{0.4724}} & 
            0.4310\\
            & AR5\_{200}   &    0.6964   &      0.6731    &    0.5723 &
            0.5533\\
            & AR5\_{All}  & \textbf{\textit{0.5516}}      &     \textbf{\textit{0.5110}}   &    0.4736  &
            0.4454\\
            & ETS\_{200} &  0.5525    &      0.5286    &     0.5025 &
            0.4847\\
            & ETS\_{All} & 0.6118    & 0.5305    &   0.4883  &  \textbf{\textit{0.4307}}\\
            \hline
            & EXP\_{200} &     0.8409 &       0.7221 &    0.6830 &      0.5844 \\
            & EXP\_{All} &     \textbf{\textit{0.7379}} &       \textbf{\textit{0.5490}} &    \textbf{\textit{0.5939}} &      \textbf{\textit{0.4441}} \\
            & Linear\_{200} &     0.7631 &       0.6268 &    0.6137 &      0.5030 \\
            & Linear\_{All} &     0.8113 &       0.6518 &    0.6431 &      0.5102 \\
            & Plain\_{200} &     0.7838 &       0.6402 &    0.6273 &      0.5109 \\
            & Plain\_{All} &     0.9242 &       0.7768 &    0.7334 &      0.6113 \\
            \hline
            & GDW &     \textbf{0.4056} &       \textbf{0.2899} &    \textbf{0.3044} &      \textbf{0.2288} \\
            & ECW &     0.5812 &       0.4162 &    0.4598 &      0.3246 \\
			\bottomrule
			\addlinespace
			\multicolumn{5}{l}{\bf Incremental} \\
			\addlinespace
			 & AR3\_{200}   &    0.5796  &   0.5428   &   0.5001  &
            0.4538\\
            & AR3\_{All}  &      \textbf{\textit{0.5602}} & \textbf{\textit{0.5266}}      & \textbf{\textit{0.4501}}   &  \textbf{\textit{0.4281}}\\
            & AR5\_{200}  &    0.6124  & 0.5977   &   0.5377  &
            0.5142\\
            & AR5\_{All} &  0.5637   &  0.5472   &  0.4551  &
            0.4453\\
            & ETS\_{200} & 0.5874   & 0.5536    &   0.5544 &
            0.5368\\
            & ETS\_{All} & 0.5746   & 0.5424 &   0.4634   &
            0.4400\\
            \hline
            & EXP\_{200} &     0.7524 &       0.7183 &    0.6101 &      0.5821 \\
            & EXP\_{All} &     \textbf{\textit{0.6236}} &       \textbf{\textit{0.5548}} &    \textbf{\textit{0.5018}} &      \textbf{\textit{0.4480}} \\
            & Linear\_{200} &     0.6745 &       0.6245 &    0.5403 &      0.5004 \\
            & Linear\_{All} &     0.7365 &       0.6569 &    0.5836 &      0.5179 \\
            & Plain\_{200} &     0.6943 &       0.6392 &    0.5541 &      0.5098 \\
            & Plain\_{All} &     0.8710 &       0.7848 &    0.6892 &      0.6175 \\
            \hline
             & GDW &     \textbf{0.3629} &       \textbf{0.3083} &    \textbf{0.2805} &      \textbf{0.2437} \\
            & ECW &     0.5541 &       0.4420 &    0.4320 &      0.3444 \\
			\bottomrule
			\addlinespace
			\multicolumn{5}{l}{\bf Gradual} \\
			\addlinespace
			  & AR3\_{200} &      0.7931 &  0.7877  & 0.7588  &
            0.7480\\
            & AR3\_{All} &    0.7939   &  0.7903  & \textbf{\textit{0.6819}}  & 
            \textbf{\textit{0.6832}}\\
            & AR5\_{200} &    0.8011   & 0.7957  & 0.7832   &
            0.7795\\
            & AR5\_{All} &   0.8637   &  0.8610  &  0.8531 &
            0.8499\\
            & ETS\_{200} &  \textbf{\textit{0.7853}}   &   \textbf{\textit{0.7810}}  & 0.6973  &
            0.6926\\
            & ETS\_{All} & 0.7936  & 0.7924  & 0.7328   &
            0.7297\\
            \hline
            & EXP\_{200} &     1.0023 &       0.9963 &    0.7810 &      0.7745 \\
            & EXP\_{All} &     1.0292 &       1.0260 &    0.7825 &      0.7821 \\
            & Linear\_{200} &     \textbf{\textit{0.8998}} &       \textbf{\textit{0.9006}} &    \textbf{\textit{0.6962}} &      \textbf{\textit{0.6961}} \\
            & Linear\_{All} &     1.1786 &       1.1469 &    0.9170 &      0.8921 \\
            & Plain\_{200} &     0.9067 &       0.9058 &    0.7036 &      0.7024 \\
            & Plain\_{All} &     1.3042 &       1.2433 &    1.0200 &      0.9722 \\
            \hline
            & GDW &     \textbf{0.7168} &       \textbf{0.7161} &    \textbf{0.4617} &      \textbf{0.4605} \\
            & ECW &     0.7716 &       0.7711 &    0.5686 &      0.5674 \\
			\bottomrule
		\end{tabular*}
		\label{tab:all_results}
\end{table}

The models in Table \ref{tab:all_results} are grouped based
on the sub-experiments. The results of the best performing models in each group are italicized, and the overall best performing models across the datasets are highlighted in boldface. The first group contains the statistical baselines we considered which are AR(3), AR(5) and ETS. The second group contains the LightGBM baselines we considered which are the plain LightGBM model (Plain), LightGBM trained with exponential weighting (EXP) and LightGBM trained with linear weighting (Linear). The baseline models also indicate the number of data points used for model training, which is either last 200 data points or all data points. The third group contains our proposed concept drift handling methods, ECW and GDW, which use two LightGBM models to provide forecasts. The hyperparameters of all LightGBM models such as maximum depth, number of leaves, minimum number of instances in a leaf, learning rate and sub-feature leaf are tuned using a grid search approach. For our experiments, we use the Python \verb|lightgbm| package. 

In the first group, different statistical models show the best performance across different concept drift types. Overall, AR5\_All and AR3\_All respectively show the best performance across sudden and incremental concept drift types. Across gradual concept drift, ETS\_200 and AR3\_All respectively show the best performance on RMSE and MAE metrics.

In the second group, EXP\_All shows the best performance across sudden and incremental concept drift types whereas Linear\_200 shows the best performance across gradual concept drift. The models that use exponential weighting overall show a better performance than the models that use linear weighting. Across all concept drift types, overall, the methods that use exponential and linear weighting show a better performance than the plain LightGBM models which do not handle concept drift. This shows assigning higher weights for the recent series observations is beneficial in handling concept drift. For different concept drift types, usage of different numbers of training data points shows best results and this indicates using two models trained with the recent series history and full series history may be a better way to handle concept drift.

In the third group, our proposed GDW method shows the best performance across all concept drift types. In particular, this method outperforms all considered baselines and shows the best performance on all error metrics. Our proposed ECW method also outperforms all considered baselines on all error metrics across all concept drift types except for 2 cases: AR5\_All and ETS\_200 on mean RMSE across sudden concept drift. This further shows combining the forecasts of two models trained with the most recent series history and full series history is a proper way of handling concept drift in global time series forecasting. The model trained with the full series history has a high forecasting accuracy where the model trained with the recent series history can quickly adapt to new concepts. Thus, combining the forecasts provided by these two models leads to high forecasting accuracy in particular for the series showing concept drift. 

The performance improvements gained by the GDW method across different concept drift types are relative to the characteristics of the drift type. In particular, compared to the LightGBM baselines, the GDW method shows considerable improvements across the sudden and incremental concept drift types and smaller improvements across the gradual concept drift type in terms of mean RMSE and mean MAE. The drift occurs gradually in the series showing gradual concept drift and thus, there is a possibility that the models have seen instances of the new distribution during training. Hence, the baseline LightGBM models have also provided accurate forecasts up to some extent for the series showing gradual concept drift. However, with the series showing sudden or incremental concept drift, there is a high possibility that the models have seen fewer or no instances of the new distribution during training where the drift should be properly handled by the models to provide accurate forecasts. Our proposed GDW method can properly handle concept drift compared to the baselines and thus, across sudden and incremental concept drift types, it shows considerable performance improvements in terms of the forecasting accuracy.  

\subsection{Statistical Testing Results}

Table \ref{tab:statistical_testing} shows the results of the statistical testing evaluation, namely the adjusted $p$-values calculated from the Friedman test with Hochberg's post-hoc procedure considering a significance level of $\alpha = 0.05$ \citep{garcia2010advanced}. The statistical testing is separately conducted using the three experimental datasets corresponding with sudden, incremental and gradual concept drift types. For a given dataset, the RMSE values of each series provided by all methods are considered. 

\begin{table}[h]
    \caption{Results of statistical testing across sudden, incremental and gradual concept drift types.}
    \begin{subtable}[h]{0.3\textwidth}
        \centering
        \caption{Sudden concept drift}
        \begin{tabular}{l|c}
        \hline
        \textbf{Method} & \textbf{$p_{\textit{Hoch}}$}  \\
        \hline
            GDW & -- \\
            \hline
            ECW &  $3.56\times 10^{-9}$ \\
            EXP\_{All} &  $< 10^{-30}$ \\
            Linear\_{200} &  $< 10^{-30}$ \\
            Plain\_{200} &  $< 10^{-30}$ \\
            Linear\_{All} & $< 10^{-30}$ \\
            EXP\_{200} & $< 10^{-30}$ \\
            Plain\_{All} & $< 10^{-30}$\\
            AR3\_{200} & $< 10^{-30}$ \\
            AR3\_{All} & $< 10^{-30}$ \\
            AR5\_{200} & $< 10^{-30}$ \\
            AR5\_{All} & $< 10^{-30}$ \\
            ETS\_{200} & $< 10^{-30}$ \\
            ETS\_{All} & $< 10^{-30}$ \\
        \hline
    \end{tabular}
    \label{tab:sudden_average_fri_pvalue_RMSE}
    \end{subtable}
    \hfill
    \begin{subtable}[h]{0.3\textwidth}
        \centering
        \caption{Incremental concept drift}
        \begin{tabular}{l | c}
        \hline
        \textbf{Method} & \textbf{$p_{\textit{Hoch}}$}  \\
        \hline
            GDW & -- \\
            \hline
            ECW & $ 3.48\times 10^{-20}$\\
            EXP\_{All} & $< 10^{-30}$\\
            Linear\_{200} & $< 10^{-30}$\\
            Plain\_{200} & $< 10^{-30}$\\
            EXP\_{200} & $< 10^{-30}$\\
            Linear\_{All} & $< 10^{-30}$\\
            Plain\_{All} & $< 10^{-30}$\\
            AR3\_{200} & $< 10^{-30}$ \\
            AR3\_{All} & $< 10^{-30}$ \\
            AR5\_{200} & $< 10^{-30}$ \\
            AR5\_{All} & $< 10^{-30}$ \\
            ETS\_{200} & $< 10^{-30}$ \\
            ETS\_{All} & $< 10^{-30}$ \\
        \hline
    \end{tabular}
    \label{tab:incremental__average_fri_pvalue_RMSE}
     \end{subtable}
     \hfill
    \begin{subtable}[h]{0.3\textwidth}
        \centering
        \caption{Gradual concept drift}
        \begin{tabular}{l | c}
        \hline
        \textbf{Method} & \textbf{$p_{\textit{Hoch}}$}  \\
        \hline
            GDW & -- \\
            \hline
            ECW &  $8.01\times 10^{-8}$\\
            Linear\_{200} &  $< 10^{-30}$\\
            Plain\_{200} &  $< 10^{-30}$\\
            EXP\_{200} & $< 10^{-30}$\\
            EXP\_{All} & $< 10^{-30}$\\
            Linear\_{All} & $< 10^{-30}$\\
            Plain\_{All} &  $< 10^{-30}$\\   
            AR3\_{200} & $< 10^{-30}$ \\
            AR3\_{All} & $< 10^{-30}$ \\
            AR5\_{200} & $< 10^{-30}$ \\
            AR5\_{All} & $< 10^{-30}$ \\
            ETS\_{200} & $< 10^{-30}$ \\
            ETS\_{All} & $< 10^{-30}$ \\
        \hline
    \end{tabular}
    \label{tab:gradual__average_fri_pvalue_RMSE}
     \end{subtable}
     \label{tab:statistical_testing}
\end{table}

The overall $p$-values of the Friedman rank-sum test corresponding with the sudden, incremental and gradual concept drift types are less than $10^{-30}$ showing the results are highly significant. For all concept drift types, GDW performs the best on ranking over RMSE per each series and thus, it is used as the control method as mentioned in the first row in each sub-table of Table \ref{tab:statistical_testing}. In all sub-tables, horizontal lines are used to separate the models that perform significantly worse than GDW. All benchmarks and ECW show $p_{\textit{Hoch}}$ values less than $\alpha$ across all concept drift types and thus, they are significantly worse than the control method. 

\subsection{Further Insights}

\begin{figure}[!htbp]
    \centering
    \includegraphics[scale=0.38]{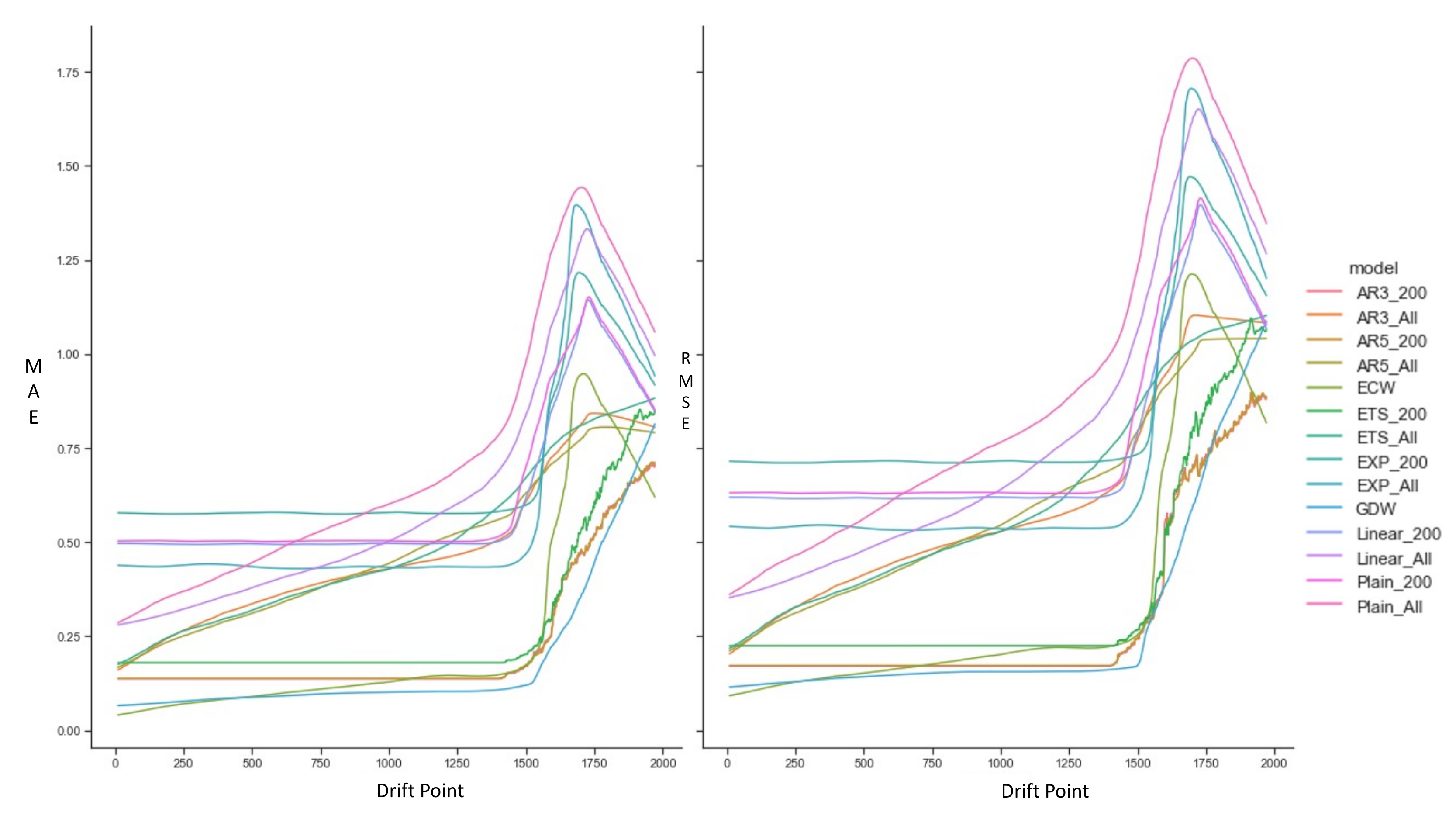}
    \caption{Performance of all models in terms of MAE (left) and RMSE (right), across the series showing sudden concept drift with different drift points.}
    \label{fig:sudden_per_adapt}
\end{figure}

Figure \ref{fig:sudden_per_adapt} shows the performance of all models in terms of MAE and RMSE, across the series showing sudden concept drift with different drift points. According to the figure, our proposed methods, GDW and ECW, show a better performance compared to the baselines in both RMSE and MAE where the GDW shows the best performance in both error metrics. Compared to the ECW and other baselines, the GDW shows a lower error increasing rate. All methods show higher errors for the series where the concept drift occurs in the test set (after 1650 data points) as the models have not seen the corresponding data distributions in the training set. However, even in this phenomenon, we see that the error increasing rate of the GDW method is considerably lower compared with the other methods. 

\begin{figure}[!htbp]
    \centering
    \includegraphics[scale=0.38]{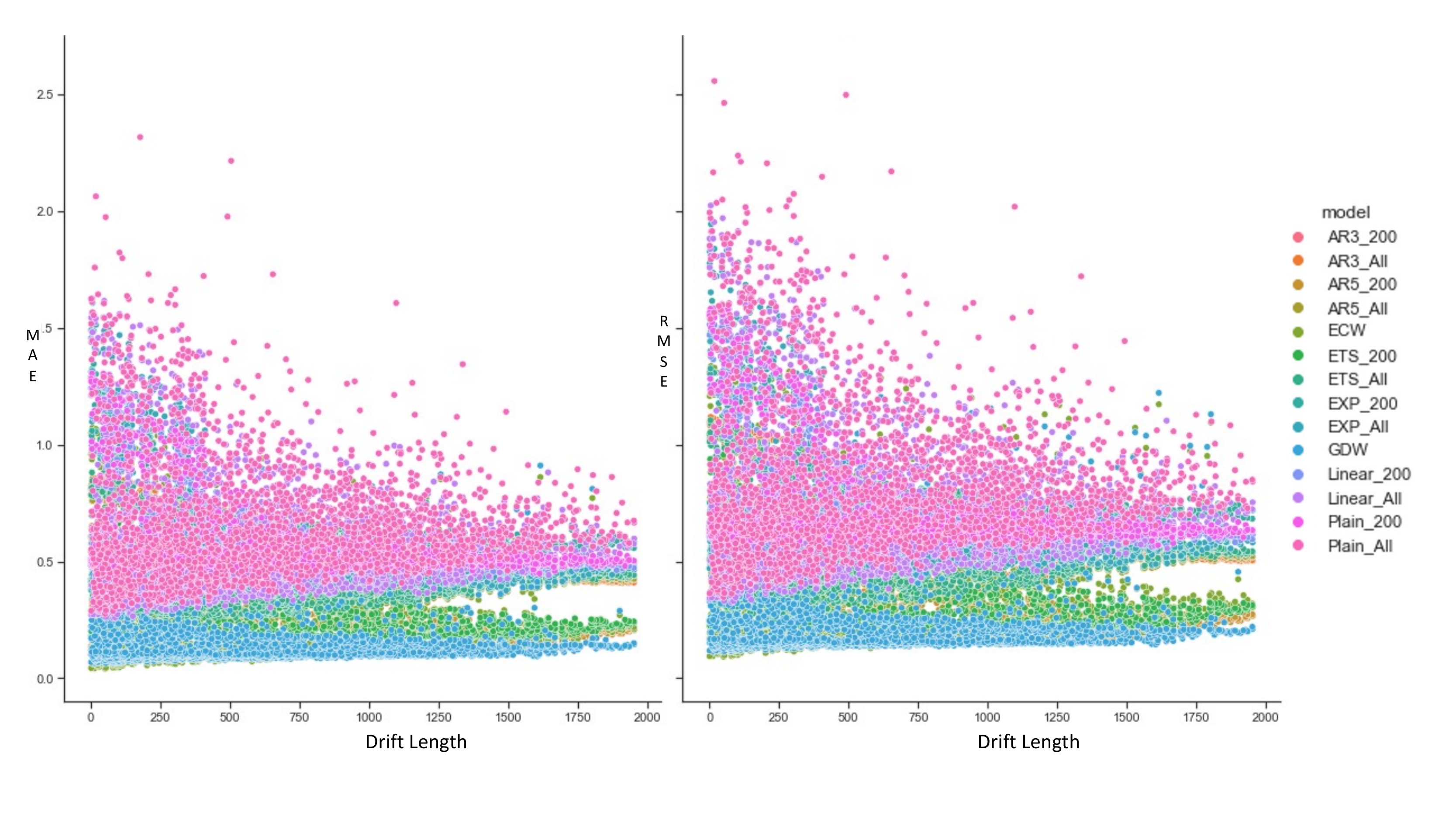}
    \caption{Performance of all models in terms of MAE (left) and RMSE (right), across the series showing incremental concept drift with different drift lengths.}
    \label{fig:incremental_per_adapt}
\end{figure}

Figure \ref{fig:incremental_per_adapt} shows the performance of all models in terms of MAE and RMSE, across the series showing incremental concept drift with different drift lengths. Here, the drift length refers to the difference between the drift starting and ending points. According to the figure, we see that all models show higher forecasting errors for shorter drift lengths, as the models are required to adapt to the new concept as quickly as possible for shorter drift lengths. Our proposed methods, GDW and ECW, are continuous adaptive weighting methods and they continuously adjust model weights based on the previous prediction error. Thus, GDW and ECW show a better performance compared to the other methods where GDW shows the best performance in both MAE and RMSE.  

\section{Conclusions and Future Research}
\label{sec:concluson}

Time series data distributions are often not stationary and as a result, the accuracy of ML forecasting models can decrease over time. Thus, handling such concept drift is a crucial issue in the field of ML methods for time series forecasting. However, the concept drift handling methods in the literature are mostly designed for classification tasks, not for forecasting. 

In this study, we have proposed two new methods based on continuous adapting weighting, GDW and ECW, to handle concept drift in global time series forecasting. Each proposed method uses two sub-models, a model trained with the most recent series history and a model trained with the full series history, where the weighted average of the forecasts provided by the two models are considered as the final forecasts. The proposed methods calculate the weights that are used to combine the sub-model forecasts based on the previous prediction error. During training, the sub-models internally assign higher weights for the recent observations either exponentially or linearly. Across three simulated datasets that demonstrate sudden, incremental and gradual concept drift types, we have shown that our proposed methods can significantly outperform a set of baseline models. In particular, GDW shows the best performance across all considered concept drift types. 

From our experiments, we conclude that using two models trained with training sets of different sizes is a proper approach to handle concept drift in global time series forecasting. We further conclude gradient descent is a proper approach to determine the weights that should be used to combine the forecasts provided by these models. Thus, we recommend the GDW method to handle concept drift in global time series forecasting.

The success of this approach encourages as future work to explore the performance of GDW and ECW methods when using three or more sub-models trained using different numbers of data points. Analysing the performance of the proposed methods based on the characteristics of the datasets such as the number of series, and the order and parameters of the AR data generation process is also worthwhile to study. It will be also interesting to study the performance of our proposed methods on real-world datasets that adequately demonstrate concept drift. Analysing the performance of our proposed methods across different base learners including deep learning models is also a worthwhile endeavour. 

%
%
\bibliographystyle{unsrtnat}
\bibliography{sample}

\begin{thebibliography}{37}
\providecommand{\natexlab}[1]{#1}
\providecommand{\url}[1]{\texttt{#1}}
\expandafter\ifx\csname urlstyle\endcsname\relax
  \providecommand{\doi}[1]{doi: #1}\else
  \providecommand{\doi}{doi: \begingroup \urlstyle{rm}\Url}\fi

\bibitem[Hyndman et~al.(2008)Hyndman, Koehler, Ord, and Snyder]{ref_112}
R.~J. Hyndman, A.~B. Koehler, J.~K. Ord, and R.~D. Snyder.
\newblock \emph{Forecasting with Exponential Smoothing: The State Space
  Approach}.
\newblock Springer Science and Business Media, 2008.

\bibitem[Januschowski et~al.(2020)Januschowski, Gasthaus, Wang, Salinas,
  Flunkert, Bohlke-Schneider, and Callot]{ref_105}
T.~Januschowski, J.~Gasthaus, Y.~Wang, D.~Salinas, V.~Flunkert,
  M.~Bohlke-Schneider, and L.~Callot.
\newblock Criteria for classifying forecasting methods.
\newblock \emph{International Journal of Forecasting}, 36\penalty0
  (1):\penalty0 167--177, 2020.

\bibitem[Makridakis et~al.(2018)Makridakis, Spiliotis, and
  Assimakopoulos]{ref_30}
S.~Makridakis, E.~Spiliotis, and V.~Assimakopoulos.
\newblock The {M4} competition: Results, findings, conclusion and way forward.
\newblock \emph{International Journal of Forecasting}, 34\penalty0
  (4):\penalty0 802--808, 2018.

\bibitem[Makridakis et~al.(2022)Makridakis, Spiliotis, and
  Assimakopoulos]{makridakis_2020_m5}
S.~Makridakis, E.~Spiliotis, and V.~Assimakopoulos.
\newblock The {M5} accuracy competition: Results, findings and conclusions.
\newblock \emph{International Journal of Forecasting}, 38\penalty0
  (4):\penalty0 1346--1364, 2022.

\bibitem[Montero-Manso and Hyndman(2021)]{pablo_2020_principles}
P.~Montero-Manso and R.~J. Hyndman.
\newblock Principles and algorithms for forecasting groups of time series:
  Locality and globality.
\newblock \emph{International Journal of Forecasting}, 2021.
\newblock ISSN 0169-2070.

\bibitem[Webb et~al.(2016)Webb, Hyde, Cao, Nguyen, and
  Petitjean]{webb2016characterizing}
G.~I. Webb, R.~Hyde, H.~Cao, H.~L. Nguyen, and F.~Petitjean.
\newblock Characterizing concept drift.
\newblock \emph{Data Mining and Knowledge Discovery}, 30\penalty0 (4):\penalty0
  964--994, 2016.

\bibitem[Kolter and Maloof(2007)]{kolter2007dynamic}
J.~Z. Kolter and M.~A. Maloof.
\newblock Dynamic weighted majority: An ensemble method for drifting concepts.
\newblock \emph{The Journal of Machine Learning Research}, 8:\penalty0
  2755--2790, 2007.

\bibitem[Gomes and Enembreck(2013)]{gomes2013sae}
H.~M. Gomes and F.~Enembreck.
\newblock {SAE}: Social adaptive ensemble classifier for data streams.
\newblock In \emph{IEEE Symposium on Computational Intelligence and Data
  Mining}, pages 199--206, 2013.

\bibitem[Baier et~al.(2020)Baier, Hofmann, K{\"u}hl, Mohr, and
  Satzger]{baier2020handling}
L.~Baier, M.~Hofmann, N.~K{\"u}hl, M.~Mohr, and G.~Satzger.
\newblock Handling concept drifts in regression problems--the error
  intersection approach.
\newblock In \emph{15th International Conference on Wirtschaftsinformatik},
  2020.

\bibitem[Gomes et~al.(2017)Gomes, Barddal, Enembreck, and
  Bifet]{gomes2017survey}
H.~M. Gomes, J.~P. Barddal, F.~Enembreck, and A.~Bifet.
\newblock A survey on ensemble learning for data stream classification.
\newblock \emph{ACM Computing Surveys (CSUR)}, 50\penalty0 (2):\penalty0 1--36,
  2017.

\bibitem[Ke et~al.(2017)Ke, Meng, Finley, Wang, Chen, Ma, Ye, and
  Liu]{guolin_lightgbm_2017}
G.~Ke, Q.~Meng, T.~Finley, T.~Wang, W.~Chen, W.~Ma, Q.~Ye, and T.~Liu.
\newblock {LightGBM}: A highly efficient gradient boosting decision tree.
\newblock In \emph{Proceedings of the 31st International Conference on Neural
  Information Processing Systems}, NIPS'17, page 3149–3157, Red Hook, NY,
  USA, 2017. Curran Associates Inc.

\bibitem[Delany et~al.(2004)Delany, Cunningham, Tsymbal, and
  Coyle]{delany2004case}
S.~J. Delany, P.~Cunningham, A.~Tsymbal, and L.~Coyle.
\newblock A case-based technique for tracking concept drift in spam filtering.
\newblock In \emph{International Conference on Innovative Techniques and
  Applications of Artificial Intelligence}, pages 3--16. Springer, 2004.

\bibitem[Hyndman and Athanasopoulos(2018)]{ref_4}
R.~J. Hyndman and G.~Athanasopoulos.
\newblock \emph{{Forecasting: Principles and Practice}}.
\newblock OTexts, 2nd edition, 2018.

\bibitem[Salinas et~al.(2020)Salinas, Flunkert, Gasthaus, and
  Januschowski]{ref_99}
D.~Salinas, V.~Flunkert, J.~Gasthaus, and T.~Januschowski.
\newblock {DeepAR}: Probabilistic forecasting with autoregressive recurrent
  networks.
\newblock \emph{International Journal of Forecasting}, 36\penalty0
  (3):\penalty0 1181--1191, 2020.

\bibitem[Smyl(2020)]{ref_1}
S.~Smyl.
\newblock A hybrid method of exponential smoothing and recurrent neural
  networks for time series forecasting.
\newblock \emph{International Journal of Forecasting}, 36\penalty0
  (1):\penalty0 75--85, 2020.

\bibitem[Bandara et~al.(2020)Bandara, Bergmeir, and Smyl]{ref_2}
K.~Bandara, C.~Bergmeir, and S.~Smyl.
\newblock Forecasting across time series databases using recurrent neural
  networks on groups of similar series: A clustering approach.
\newblock \emph{Expert Systems with Applications}, 140:\penalty0 112896, 2020.

\bibitem[Hewamalage et~al.(2020)Hewamalage, Bergmeir, and Bandara]{ref_6}
H.~Hewamalage, C.~Bergmeir, and K.~Bandara.
\newblock Recurrent neural networks for time series forecasting: Current status
  and future directions.
\newblock \emph{International Journal of Forecasting}, 2020.

\bibitem[Godahewa et~al.(2021)Godahewa, Bandara, Webb, Smyl, and
  Bergmeir]{godahewa_2021_ensembling}
R.~Godahewa, K.~Bandara, G.~I. Webb, S.~Smyl, and C.~Bergmeir.
\newblock Ensembles of localised models for time series forecasting.
\newblock \emph{Knowledge-Based Systems}, 233:\penalty0 107518, 2021.

\bibitem[Oreshkin et~al.(2021)Oreshkin, Carpov, Chapados, and
  Bengio]{oreshkin_2021_meta}
B.~N. Oreshkin, D.~Carpov, N.~Chapados, and Y.~Bengio.
\newblock Meta-learning framework with applications to zero-shot time-series
  forecasting.
\newblock In \emph{AAAI Conference on Artificial Intelligence}, volume~35,
  2021.

\bibitem[Mergenthaler and Ramírez(2022)]{mergenthaler_2022_nixtla}
M.~Mergenthaler and F.~G. Ramírez.
\newblock Nixtla: Transfer learning for time series forecasting, 2022.
\newblock URL \url{https://github.com/Nixtla/transfer-learning-time-series}.

\bibitem[Oreshkin et~al.(2020)Oreshkin, Carpov, Chapados, and
  Bengio]{oreshkin_2019_nbeats}
B.~N. Oreshkin, D.~Carpov, N.~Chapados, and Y.~Bengio.
\newblock {N-BEATS}: Neural basis expansion analysis for interpretable time
  series forecasting.
\newblock In \emph{8th International Conference on Learning Representations
  {(ICLR)}}, 2020.

\bibitem[Challu et~al.(2023)Challu, Olivares, Oreshkin, Garza,
  Mergenthaler-Canseco, and Dubrawski]{challu_2023_nhits}
C.~Challu, K.~G. Olivares, B.~N. Oreshkin, F.~Garza, M.~Mergenthaler-Canseco,
  and A.~Dubrawski.
\newblock {N-HiTS}: Neural hierarchical interpolation for time series
  forecasting.
\newblock In \emph{AAAI Conference on Artificial Intelligence}, volume~37,
  2023.

\bibitem[Woo et~al.(2022)Woo, Liu, Sahoo, Kumar, and Hoi]{woo_2022_deeptime}
G.~Woo, C.~Liu, D.~Sahoo, A.~Kumar, and S.~Hoi.
\newblock Deeptime: Deep time-index meta-learning for non-stationary
  time-series forecasting.
\newblock \url{https://arxiv.org/abs/2207.06046}, 2022.

\bibitem[Grazzi et~al.(2021)Grazzi, Flunkert, Salinas, Januschowski, Seeger,
  and Archambeau]{grazzi_2021_meta}
R.~Grazzi, V.~Flunkert, D.~Salinas, T.~Januschowski, M.~Seeger, and
  C.~Archambeau.
\newblock Meta-forecasting by combining global deep representations with local
  adaptation.
\newblock \url{https://arxiv.org/abs/2111.03418}, 2021.

\bibitem[Ye and Dai(2018)]{ref_106}
R.~Ye and Q.~Dai.
\newblock A novel transfer learning framework for time series forecasting.
\newblock \emph{Knowledge-Based Systems}, 156:\penalty0 74--99, 2018.

\bibitem[Ghomeshi et~al.(2019)Ghomeshi, Gaber, and Kovalchuk]{ghomeshi2019eacd}
H.~Ghomeshi, M.~M. Gaber, and Y.~Kovalchuk.
\newblock {EACD}: Evolutionary adaptation to concept drifts in data streams.
\newblock \emph{Data Mining and Knowledge Discovery}, 33\penalty0 (3):\penalty0
  663--694, 2019.

\bibitem[Chu and Zaniolo(2004)]{chu2004fast}
F.~Chu and C.~Zaniolo.
\newblock Fast and light boosting for adaptive mining of data streams.
\newblock In \emph{Pacific-Asia Conference on Knowledge Discovery and Data
  Mining}, pages 282--292. Springer, 2004.

\bibitem[Krawczyk et~al.(2017)Krawczyk, Minku, Gama, Stefanowski, and
  Wo{\'z}niak]{krawczyk2017ensemble}
B.~Krawczyk, L.~Minku, J.~Gama, J.~Stefanowski, and M.~Wo{\'z}niak.
\newblock Ensemble learning for data stream analysis: A survey.
\newblock \emph{Information Fusion}, 37:\penalty0 132--156, 2017.

\bibitem[Gon{\c{c}}alves~Jr and De~Barros(2013)]{gonccalves2013rcd}
P.~M. Gon{\c{c}}alves~Jr and R.~S.~S. De~Barros.
\newblock {RCD}: A recurring concept drift framework.
\newblock \emph{Pattern Recognition Letters}, 34\penalty0 (9):\penalty0
  1018--1025, 2013.

\bibitem[Widmer and Kubat(1996)]{widmer1996learning}
G.~Widmer and M.~Kubat.
\newblock Learning in the presence of concept drift and hidden contexts.
\newblock \emph{Machine Learning}, 23\penalty0 (1):\penalty0 69--101, 1996.

\bibitem[Januschowski et~al.(2021)Januschowski, Wang, Torkkola, Erkkilä,
  Hasson, and Gasthaus]{JANUSCHOWSKI2021}
T.~Januschowski, Y.~Wang, K.~Torkkola, T.~Erkkilä, H.~Hasson, and J.~Gasthaus.
\newblock Forecasting with trees.
\newblock \emph{International Journal of Forecasting}, 2021.

\bibitem[sam(2010)]{sammut_2010_mae}
Mean absolute error.
\newblock In C.~Sammut and G.~I. Webb, editors, \emph{Encyclopedia of Machine
  Learning}, pages 652--652. Springer US, Boston, MA, 2010.

\bibitem[Dawid(1984)]{dawid1984present}
A.~P. Dawid.
\newblock Present position and potential developments: Some personal views
  statistical theory the prequential approach.
\newblock \emph{Journal of the Royal Statistical Society: Series A (General)},
  147\penalty0 (2):\penalty0 278--290, 1984.

\bibitem[Cerqueira et~al.(2020)Cerqueira, Torgo, and
  Mozeti{\v{c}}]{cerqueira2020evaluating}
V.~Cerqueira, L.~Torgo, and I.~Mozeti{\v{c}}.
\newblock Evaluating time series forecasting models: An empirical study on
  performance estimation methods.
\newblock \emph{Machine Learning}, 109\penalty0 (11):\penalty0 1997--2028,
  2020.

\bibitem[Tashman(2000)]{tashman2000out}
L.~J. Tashman.
\newblock Out-of-sample tests of forecasting accuracy: An analysis and review.
\newblock \emph{International journal of forecasting}, 16\penalty0
  (4):\penalty0 437--450, 2000.

\bibitem[Garza et~al.(2022)Garza, Canseco, Challú, and
  Olivares]{garza2022statsforecast}
F.~Garza, M.~M. Canseco, C.~Challú, and K.~G. Olivares.
\newblock {StatsForecast}: Lightning fast forecasting with statistical and
  econometric models.
\newblock {PyCon} Salt Lake City, Utah, US, 2022.
\newblock URL \url{https://github.com/Nixtla/statsforecast}.

\bibitem[Garc{\'\i}a et~al.(2010)Garc{\'\i}a, Fern{\'a}ndez, Luengo, and
  Herrera]{garcia2010advanced}
S.~Garc{\'\i}a, A.~Fern{\'a}ndez, J.~Luengo, and F.~Herrera.
\newblock Advanced nonparametric tests for multiple comparisons in the design
  of experiments in computational intelligence and data mining: Experimental
  analysis of power.
\newblock \emph{Information Sciences}, 180\penalty0 (10):\penalty0 2044--2064,
  2010.

\end{thebibliography}

\end{document}